\documentclass[acmsmall,screen,nonacm]{acmart}

\AtBeginDocument{%
  }

\usepackage{iftex}
\ifboolexpr{bool{xetex} or bool{luatex}}{
  \usepackage{fontspec}
  \newfontfamily\tibetanfont{NotoSerifTibetan}[Path=./fonts/,Extension=.ttf]
}{}
\newcommand{\tib}[1]{{\tibetanfont #1}}

\begin{document}

\title{Syllable- and Word-Level Evaluation of Tibetan OCR}

\author{Jacob Moore}
\email{jmoore10@ltu.edu}
\affiliation{%
  \institution{Lawrence Technological University}
  \city{Southfield}
  \state{Michigan}
  \country{USA}
}

\author{Paula Lauren}
\email{plauren@ltu.edu}
\affiliation{%
  \institution{Lawrence Technological University}
  \city{Southfield}
  \state{Michigan}
  \country{USA}
}

\begin{abstract}
Optical character recognition (OCR) for Tibetan is evaluated predominantly by character error rate (CER). Tibetan marks syllable boundaries but does not mark word boundaries, so a word error rate requires a segmenter. This study evaluates nine recognizer configurations on typeset, woodblock and two handwritten script styles, alone and in combination, and reports syllable- and word-level error rates, to our knowledge the first for Tibetan OCR. The configurations span CTC models, TrOCR encoder--decoders with and without post-training quantization, and a vision-language model, and each is scored under CER, the syllable error rate (SER) and three segmented word error rates (SWER). CER ranks systems largely as the higher-level rates do (Kendall $\tau \geq 0.833$), with one pair ordered oppositely. On the effect of joint training, CER and SER record significant changes of opposite sign in 7 of 64 comparisons. SER needs no segmenter and ranks systems as SWER does under two of the three segmenters ($\tau \geq 0.889$). The output of each configuration is also tagged for part of speech, annotated for named entities and machine translated, and each result is compared with that obtained on the reference text. As rough guidance, translation chrF of 60 is reached at SER of 0.047 or below and part-of-speech preservation of 0.8 at 0.078 or below, while no error rate separates the systems that reach entity preservation of 0.6 from those that do not.
\end{abstract}



\maketitle

\section{Introduction}
\label{sec:intro}

Optical character recognition (OCR) and handwritten text recognition are applied to Tibetan prints and manuscripts, including to make large collections of Buddhist texts searchable \citep{luo2024cleansingjewel,griffiths2024htr,erhard2025tibetan}. Recognition of continuous Tibetan text is typically evaluated by character error rate (CER) \citep{griffiths2024htr,erhard2025tibetan,sabbagh2025htr,xu2026ftibsuite}. The orthography of Tibetan makes word error rates, typical in other languages \citep{neudecker2021survey,vesalainen2026errorpatterns}, difficult to compute.

In scripts that delimit a unit larger than the character, CER and an error rate over that unit can rank systems differently \citep{agbetimessan2026ssmocr,le2025vietnamesedar}. OCR error also degrades the downstream tasks that use the recognized text \citep{vanstrien2020impact}. Named-entity recognition falls sharply once word-level error reaches 8--20\% \citep{hamdi2019nerocr}, and corrupting named-entity spans reduces keyword search performance while CER and word error rate remain low \citep{jaud2025beyondcer}.

Tibetan explicitly marks syllable boundaries, chiefly with the tsheg (\tib{་}), but does not mark word boundaries. A word error rate therefore requires a segmenter. The syllable is an intermediate unit that requires none, and segmented word error rates have been applied to Tibetan speech recognition \citep{moore2026tibetanasr}. To our knowledge, no error rate above the character has been reported for Tibetan OCR. It is therefore not known whether CER leads an evaluator to the same choices as a syllable- or word-level rate would, or which rate best predicts the quality of downstream tasks.

Nine recognizer configurations are evaluated, spanning CTC models, TrOCR encoder--decoders with and without post-training quantization, and a vision-language model. They are trained on four datasets, covering modern typeset text, woodblock print and two handwritten styles, and on three mixed-domain combinations of them to compare joint with single-domain training. This gives 60 configuration $\times$ condition cells. Each cell is scored under CER, the syllable error rate (SER) and three segmented word error rates (SWER). On the four single-domain conditions, recognizer output is also tagged for part of speech, annotated for named entities and machine translated into English, and each result is scored against the result obtained on the reference transcription. These \emph{preservation} scores measure the effect of recognition error on each task.

This study makes three contributions.

\begin{enumerate}
\item \textbf{A comparative evaluation of Tibetan OCR across architectures, quantization and script domains.} Configurations are compared on page-grouped held-out splits with bootstrap intervals, under syllable- and word-level error rates that are, to our knowledge, the first publicly reported for Tibetan OCR. 

\item \textbf{A comparison of CER with syllable- and word-level rates.} CER ranks systems largely as the higher-level rates do (Kendall $\tau \geq 0.833$), but on the effect of joint training it records significant changes of opposite sign to SER in 7 of 64 comparisons. Word-level rankings agree across three segmenters at $\tau \geq 0.833$ (\S\ref{sec:res-agreement}, \S\ref{sec:disc-rankings}, \S\ref{sec:disc-interventions}).

\item \textbf{Approximate error rates for downstream use.} The SER needed for a rough preservation target ranges from 0.007 for entities to 0.078 for part-of-speech tags, so no single rate serves every task. LLM annotation also fails more often as OCR error rises (\S\ref{sec:res-downstream}, \S\ref{sec:disc-downstream}).
\end{enumerate}

Sections~\ref{sec:background} and~\ref{sec:methods} give the background and methods, Section~\ref{sec:results} the results and Section~\ref{sec:discussion} their interpretation. Section~\ref{sec:future} states limitations and future work, Section~\ref{sec:conclusion} concludes, and Appendix~\ref{app:prompts} gives the prompts used.

\section{Background and Prior Work}
\label{sec:background}

\subsection{Tibetan Script}
\label{sec:script}

\subsubsection{Structure}
\label{sec:script-structure}

Tibetan is written in an \emph{abugida}, a writing system whose base unit is a consonant carrying an inherent vowel, modified by diacritics. The fundamental unit of written Tibetan is the syllable, built from a root consonant that may carry superscript and subscript letters, a vowel diacritic, and prefix and suffix consonants \citep{huang2025tibetansurvey}. The root and its superscript and subscript form a single visual column, called a \emph{stack}. Unicode encodes a stack as a sequence of codepoints rather than as a unit, so one visual column may consist of several codepoints.

Syllables are separated by the tsheg (\tib{་}, U+0F0B) \citep{vanschaik2014palaeography}. Word boundaries are not marked. Nearly every word boundary is also a syllable boundary, but words may be polysyllabic \citep{brown2024tibetan}, as also in Dzongkha \citep{norbu2010dzongkhaseg}. Punctuation is sparse: the only mark occurring throughout running text is the shad (\tib{།}, U+0F0D), which separates units larger than the syllable, roughly sentences \citep{vanschaik2014palaeography}, and so also ends the syllable before it. Syllable-level tokenization therefore splits on the tsheg, the shad and whitespace. Word-level tokenization requires a more complex segmentation step.

\subsubsection{Varieties and Media}
\label{sec:script-varieties}

Written Tibetan divides into two script families, distinguished by the presence or absence of a \emph{head}, a short horizontal stroke along the top of many letters \citep{vanschaik2014palaeography}. \emph{Uchen} (`headed script') is the formal block script and the basis for woodblock-carved prints and modern digital typesetting. \emph{Ume} (`headless script') is the cursive family, whose letters lack a head and are written with more connected strokes, with substantially greater variability in letterform shape and positioning. Within Ume, \emph{Drutsa} is a handwritten cursive \citep{erhard2025tibetan}, and \emph{Betsug} (`book form') is a calligraphic style \citep{vanschaik2014palaeography}. The two families differ enough that recognition does not necessarily transfer between them \citep{griffiths2024htr}.

The physical medium varies independently of script variety. In woodblock prints each character is carved by hand, so letterforms vary more than in machine print, and manual printing can merge the strokes of neighbouring characters \citep{hedayati2010tibetan}. Modern typeset text is uniform in letterform, while handwriting varies between writers \citep{zhou2025tibhcr}.

\subsection{Recognition Architectures}
\label{sec:architectures}

\subsubsection{Recognizers}
\label{sec:ctc}

Early Tibetan OCR systems decoded with hidden Markov models \citep{hedayati2010tibetan,rowinski2016namsel}. More recent Tibetan recognition is neural, including convolutional and convolutional-recurrent networks \citep{zhang2022uchen,wang2024unsupervised,zhou2025tibhcr}, an encoder--decoder for scene text \citep{ke2025scene}, the PyLaia engine of Transkribus \citep{griffiths2024htr,erhard2025tibetan} and a vision-language model adapted to Tibetan \citep{xu2026ftibsuite}. This study compares three kinds of neural recognizer. Connectionist temporal classification \citep[CTC;][]{graves2006ctc} trains a sequence model to maximize the likelihood of the target over all valid frame-to-character alignments. Convolutional-recurrent networks \citep{shi2017crnn} and Easter2.0 \citep{chaudhary2022easter2}, which replaces the recurrence with convolutions, are both trained with CTC. Autoregressive encoder-decoders generate over the encoder output instead, as in TrOCR \citep{li2023trocr}. Vision-language models pair a vision encoder with a language decoder pretrained on a large multilingual corpus rather than on the recognition task; Qwen3-VL \citep{bai2025qwen3vl}, whose decoder is built on Qwen3 \citep{yang2025qwen3}, is one such model.

\subsubsection{Quantization and Adaptation}
\label{sec:peft}

Quantization stores weights at lower numerical precision and may be applied to an already-trained model \citep{dettmers2022llmint8}. Quantized low-rank adaptation \citep[QLoRA;][]{dettmers2023qlora} quantizes a frozen base model and trains only small low-rank adapter matrices \citep{hu2022lora}, which reduces the memory needed to fine-tune a large model.

\subsection{Edit-Distance Error Rates}
\label{sec:metrics}

In a full OCR pipeline, layout analysis and line segmentation precede recognition. This study takes line images as given and evaluates recognition alone. The recognized text of each line is scored against a reference transcription by the number of edits needed to turn one into the other, relative to the length of the reference.

Character, syllable and word error rates share the same formula,
\[
\mathrm{ER} = \frac{S + D + I}{N},
\]
where $S$, $D$ and $I$ are the substitutions, deletions and insertions in the \emph{hypothesis} with respect to the \emph{reference}, and $N$ is the count of units in the reference. \emph{Character error rate} (CER) tokenizes over Unicode characters, \emph{syllable error rate} (SER) over syllables delimited by the tsheg, the shad or whitespace, and \emph{segmented word error rate} (SWER) over words, which for Tibetan requires a segmenter (\S\ref{sec:script-structure}). Three SWER variants were introduced for Tibetan automatic speech recognition (ASR) by \citet{moore2026tibetanasr}. BoTok-SWER segments with the dictionary-based \texttt{botok} tokenizer, BERT-SWER with a Tibetan BERT token classifier (\texttt{tibetan-bert-base-upos}), and Gem-SWER with Gemini 2.5 Flash Lite, prompted at temperature 0.0. Their agreement with human word-boundary annotation on speech transcripts, as reported there, is Cohen's $\kappa$ = 0.027, 0.372 and 0.656 respectively. For all of the error rates described here, lower error-rate values indicate better performance. Values above 1.0 occur when the edits outnumber the reference units, which requires insertions.

Because the three levels differ only in tokenization, they are not independent measurements, but errors falling on unit boundaries affect them differently: a single inserted or deleted delimiter counts as one character edit but corrupts both units it separates.

\subsection{Prior Work}
\label{sec:prior-work}

\subsubsection{Evaluation Practice in Tibetan OCR}
\label{sec:eval-practice}

Early Tibetan OCR systems scored output in units of Tibetan orthography rather than codepoints. \citet{hedayati2010tibetan} evaluate a woodblock recognizer by Levenshtein distance over stack characters rather than codepoints, and \citet{rowinski2016namsel} reports isolated-character accuracy and scores each page by a syllable validity ratio, the proportion of syllables that follow Tibetan spelling rules or appear in a list of accepted exceptions.

Recent OCR and handwritten text recognition (HTR) on continuous Tibetan text typically reports CER. Much of this work is built on Transkribus, which returns CER \citep{griffiths2024htr,erhard2025tibetan}, and \citet{sabbagh2025htr} reports mean and median CER on a held-out set. FTibSuite \citep{xu2026ftibsuite} pairs CER with Exact Match, a whole-line binary, and notes that changes in the two are not perfectly aligned. Other tasks use other measures: work on isolated characters reports classification accuracy \citetext{\citealp{wang2024unsupervised,zhang2022uchen}; see also the survey by \citealp{gongqu2024survey}}. Work on scene text, evaluated on photographs of street views and outdoor environments, reports line accuracy, character accuracy and a normalized edit distance \citep{ke2025scene}.

To our knowledge, no syllable- or word-level error rate has been reported for Tibetan OCR, though SER has been reported in Tibetan ASR \citep{wang2020tibetanmultitask,zhao2019tibetanmultitask}.

\subsubsection{Character and Word Error Rates in Other Scripts}
\label{sec:beyond-cer}

CER requires no tokenization beyond the codepoint and is computable for any script, including those that do not delimit words with whitespace. \citet{thennal2025advocatingcer} advocate it as the preferred metric for multilingual ASR, on the grounds that for languages without word separators ``the WER can only be applied after a subjective and potentially inaccurate word segmentation process''. Similar reasoning appears in OCR. \citet{clerice2022groundtruthfree} judges WER ``complicated to apply'' to Old French and Medieval Latin, where spaces vary in size or are absent.

Where whitespace does delimit words, word error rate is reported alongside CER and comparison across the two levels is normal. The two levels often agree. \citet{neudecker2021survey} plot CER and WER side by side, WER exceeding CER throughout, and the two systems compared by \citet{vesalainen2026errorpatterns} are ranked the same way by both, though they differ in the kind of edits they make. Agreement is not general. \citet{agbetimessan2026ssmocr}, benchmarking historical newspaper OCR on Latin script, report a different best system under CER than under WER on each of their two subcorpora, though under 0.1 percentage points separates the CER of their leading models.

An intermediate unit has been used in scripts where word boundaries are ambiguous: \citet{naing2015myanmarlvcsr} propose a syllable error rate for Myanmar speech recognition. Vietnamese words often span several space-separated syllables \citep{le2025vietnamesedar}, so a word error rate over space-delimited tokens is in effect a syllable error rate. In that survey's line-level results the system with the lower CER (3.24\% against 4.67\%) has the higher syllable-level rate (14.11\% against 13.33\%).

For Tibetan, the only comparison against a word-level metric is in speech recognition. \citet{moore2026tibetanasr} benchmark five ASR architectures and six adaptation strategies under CER, SER and the three SWER variants, and report ranking differences between SER and CER, the same architectural ranking under all three word-level metrics, and a fall in the CER--SER correlation under parameter-efficient fine-tuning.

\subsubsection{Downstream Use of Recognition Output}
\label{sec:downstream-prior}

Recognition output is often the input to a further process, and its quality has been measured by the degradation it causes there. \citet{vanstrien2020impact} evaluate six tasks across OCR output of varying quality and report a consistent impact across all six. \citet{hamdi2019nerocr} find named-entity F-score falling from roughly 90\% to 60\% once word-level error rates reach the 8--20\% range. \citet{jaud2025beyondcer} report that corrupting named-entity spans sharply reduces keyword search performance while CER and WER remain low, whereas corrupting the same proportion of non-entity tokens raises both rates and leaves retrieval comparatively stable.

Downstream resources for Tibetan are limited. \citet{meelen2017tibetanpos} tag Classical Tibetan under a purpose-built tagset. A 20,215-sentence Tibetan named-entity corpus is available on request rather than publicly released, and recent entity recognition work on it fine-tunes a large language model with domain prompts \citep{zhang2025tibetanner}. Tibetan machine translation has concentrated on the Tibetan--Chinese pair \citetext{\citealp{moore2025tibetanmt}; e.g.\ \citealp{zhou2024tibetanchinesemt}}, with few studies of Tibetan--English \citep{shu2024transcending, nehrdich2026mitra, moore2025tibetanmt}.

\citet{griffiths2024htr} notes that a CER under 10\% is ``considered efficient'' for automatic transcription, and sets 5\% as the project's target for text mining, though neither threshold is explicitly validated.

\section{Methods}
\label{sec:methods}

Nine recognizer configurations are evaluated on seven data conditions, four single-domain and three mixed-domain (\S\ref{sec:datasets}--\S\ref{sec:training}). Each of the resulting 60 cells is scored on a held-out test split under five error rates (\S\ref{sec:eval-metrics}). Error concentration is measured per cell (\S\ref{sec:eval-operations}), and the rates are compared across systems, lines and training conditions (\S\ref{sec:eval-stats}). On the four single-domain conditions, recognizer output is also passed to three downstream tasks (\S\ref{sec:downstream}).

\subsection{Datasets and Splits}
\label{sec:datasets}
\label{sec:single-domain}

\begin{figure}[t]
\centering
\begin{minipage}{\linewidth}
\textbf{KhyentseWangpo} --- modern typeset\\[2pt]
\includegraphics[height=9mm]{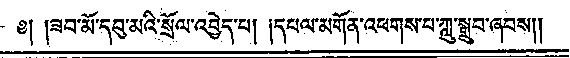}\\[3pt]
{\small\tib{༅། །ཟབ་མོ་དབུ་མའི་སྲོལ་འབྱེད་པ། །དཔལ་མགོན་འཕགས་པ་ཀླུ་སྒྲུབ་ཞབས།།}}
\end{minipage}

\vspace{6pt}

\begin{minipage}{\linewidth}
\textbf{Karmapa8} --- woodblock print\\[2pt]
\includegraphics[height=9mm]{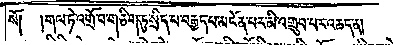}\\[3pt]
{\small\tib{སོ། །གལ་ཏེ་འགྲོ་བ་གཅིག་ཏུ་སྲིད་པ་བརྒྱད་པ་མངོན་པར་མི་འགྲུབ་པར་འཆད་ན།}}
\end{minipage}

\vspace{6pt}

\begin{minipage}{\linewidth}
\textbf{Drutsa} --- cursive handwriting\\[2pt]
\includegraphics[height=9mm]{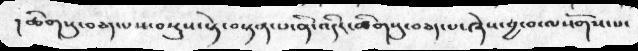}\\[3pt]
{\small\tib{།  ཚིགསུ་བཅད་པས་བསྡུས་སྟེ་བསྟན་པ་ནི་འདིར་ཚིགསུ་བཅད་པ་ཞེས་བྱ་བ་ལསོགས་པ་}}
\end{minipage}

\vspace{6pt}

\begin{minipage}{\linewidth}
\textbf{Betsug} --- semi-cursive handwriting\\[2pt]
\includegraphics[height=9mm]{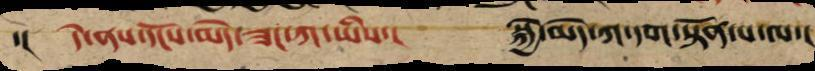}\\[3pt]
{\small\tib{་། དེ་ནས་དངོས་མེད་ཟང་ཀ་ཡིས་། སྐྱེ་མེད་ཀ་དག་སྟོན་པ་ལ་།}}
\end{minipage}

\vspace{6pt}

\caption{One test line from each script domain, with its reference
transcription.}
\label{fig:example-lines}
\Description{Four line images, one per script domain, each shown above its
reference transcription in Tibetan script.}
\end{figure}

{\sloppy Four publicly available datasets from the Hugging Face Hub are used, spanning the principal script domains of historical and contemporary Tibetan OCR (\S\ref{sec:script-varieties}): KhyentseWangpo (\texttt{BDRC/\allowbreak KhyentseWangpo}, modern typeset), Karmapa8 (\texttt{BDRC/\allowbreak Karmapa8}, woodblock print), Drutsa (\texttt{openpecha/\allowbreak OCR-Drutsa}, cursive \emph{dbu med} handwriting) and Betsug (\texttt{openpecha/\allowbreak OCR-Betsug}, semi-cursive \emph{dbu med} handwriting). All four supply pre-cropped line images paired with transcriptions. Statistics are given in Table~\ref{tab:dataset-stats} and example lines in Figure~\ref{fig:example-lines}.\par}

\begin{table}[t]
\caption{Dataset statistics. Mean chars and Vocab are computed on the training split. Identifiers refer to public Hugging Face repositories. The three mixed-domain conditions concatenate components, and their line counts are the sums of the component counts.}
\label{tab:dataset-stats}
\label{tab:mixed-domain}
\begin{tabular}{lrrrrrrrr}
\toprule
Dataset & \multicolumn{1}{c}{Total} & \multicolumn{1}{c}{Train} & \multicolumn{1}{c}{Test} & \multicolumn{1}{c}{Train} & \multicolumn{1}{c}{Test} & \multicolumn{1}{c}{Mean} & \multicolumn{1}{c}{Max} & \multicolumn{1}{c}{Vocab} \\
 & \multicolumn{1}{c}{lines} & \multicolumn{1}{c}{lines} & \multicolumn{1}{c}{lines} & \multicolumn{1}{c}{pages} & \multicolumn{1}{c}{pages} & \multicolumn{1}{c}{chars} & \multicolumn{1}{c}{chars} & \\
\midrule
KhyentseWangpo & 13{,}527 & 12{,}172 & 1{,}355 & 2{,}102 & 234 & 167 & 312 & 112 \\
Karmapa8       & 24{,}973 & 22{,}476 & 2{,}497 & 4{,}048 & 449 & 212 & 401 & 122 \\
Drutsa         & 32{,}360 & 29{,}116 & 3{,}244 & 2{,}901 & 318 & 128 & 457 & 135 \\
Betsug         & 28{,}318 & 25{,}485 & 2{,}833 & 4{,}337 & 480 & 148 & 314 & 131 \\
\midrule
Drutsa+Betsug           & 60{,}678 & 54{,}601 & 6{,}077 & & & & & \\
KhyentseWangpo+Karmapa8 & 38{,}500 & 34{,}648 & 3{,}852 & & & & & \\
All four                & 99{,}178 & 89{,}249 & 9{,}929 & & & & & \\
\bottomrule
\end{tabular}
\Description{Table of line counts, page counts, mean and maximum character lengths, and vocabulary size for the four single-domain datasets, followed by combined line counts for the three mixed-domain conditions.}
\end{table}

\label{sec:split}
All datasets are split 90/10 by page, pages drawn randomly until the test set first reaches the target size. A mixed condition is the concatenation of its components' splits, so each mixed test set is the union of the corresponding single-domain test sets.

\subsection{Models}
\label{sec:model-architectures}

Five architectures are evaluated (Table~\ref{tab:models}), spanning the CTC, autoregressive encoder-decoder and vision-language classes of \S\ref{sec:ctc}.

\begin{table}[h]
\caption{Model architectures. TrOCR checkpoints are stage-1 pretrained on synthetic printed text-line images \citep{li2023trocr}.}
\label{tab:models}
\label{tab:trocr-configs}
\label{tab:qwen-config}
\begin{tabular}{lll}
\toprule
Architecture & Type & Params \\
\midrule
CRNN        & CNN + BiLSTM, CTC            & 8.8M \\
Easter2.0   & 1D conv + SE, CTC            & 6.2M \\
TrOCR-small & DeiT-Small \citep{touvron2021deit} + 6-layer decoder & 28.7M \\
TrOCR-base  & BEiT-Base \citep{bao2022beit} + 12-layer decoder & 281.6M \\
Qwen3-VL-2B & ViT + 28-layer LLM decoder   & 2.13B \\
\bottomrule
\end{tabular}
\Description{Table of the five model architectures, their type, and parameter counts.}
\end{table}

\label{sec:ctc-models}
\paragraph{CTC models} Both follow their published specifications with two exceptions. CRNN takes input 64\,px high rather than 32\,px, because Tibetan stacks up to four glyph components vertically. Easter2.0 is reimplemented without its TACo augmentation and weighted CTC loss, so that the two CTC models share one training procedure. Line images are resized to $64\times2048$\,px, and both models produce about 512 output frames per line, more than the number of characters in any reference line.

\label{sec:trocr}
\paragraph{TrOCR} The pretrained decoders are initialized from RoBERTa and MiniLM \citep{li2023trocr}, whose subword vocabularies were not built for Tibetan. The decoder vocabulary is replaced with a Tibetan character vocabulary before fine-tuning: \texttt{embed\_tokens} and \texttt{output\_projection} are resized and randomly reinitialized, while all encoder and decoder weights are retained. Images are resized to $384\times384$ by the checkpoint's processor.

\label{sec:qwen}
\paragraph{Qwen3-VL} Its byte-level BPE covers the Tibetan block (U+0F00--0FFF) at $\sim$1.33 tokens per character. The model is prompted with a fixed instruction (Appendix~\ref{app:prompt-ocr}) and the line image. Qwen3-VL is trained only on the four single-domain datasets, with 4-bit QLoRA: adapters ($r = 8$, $\alpha = 16$, dropout $0.1$) on the attention and feed-forward projections.

\subsection{Adaptation and Training}
\label{sec:adaptation}
\label{sec:training}

Each TrOCR size is used in three configurations: standard fine-tuning, and 8-bit and 4-bit quantization (\S\ref{sec:peft}) of the fine-tuned checkpoint after training, which are not trained separately. CRNN and Easter2.0 receive standard fine-tuning only, and Qwen3-VL 4-bit QLoRA only.

\begin{table}[h]
\caption{Per-architecture hyperparameters. The 8-bit optimizer is that of \citet{dettmers2022optimizers}. \emph{Epochs} is the number of passes over the training set implied by a 4{,}000-step single-domain run at the given effective batch size, as a range over the four single-domain datasets.}
\label{tab:arch-hparams}
\label{tab:shared-hparams}
\begin{tabular}{lrrlll}
\toprule
Architecture & Train batch & Eff.\ batch & Epochs & Learning rate & Optimizer \\
\midrule
CRNN         & 16 & 128 & 17.6--42.1 & $1{\times}10^{-3}$    & AdamW \\
Easter2.0    & 16 & 128 & 17.6--42.1 & $1{\times}10^{-3}$    & AdamW \\
TrOCR-small  & 4  & 32  & 4.4--10.5  & $3.75{\times}10^{-5}$ & AdamW (8-bit) \\
TrOCR-base   & 2  & 16  & 2.2--5.3   & $2.5{\times}10^{-5}$  & AdamW (8-bit) \\
Qwen3-VL-2B  & 1  & 8   & 1.1--2.6   & $1{\times}10^{-4}$    & AdamW (8-bit) \\
\bottomrule
\end{tabular}
\Description{Table of per-architecture training batch size, effective batch size, implied epoch count, learning rate, and optimizer.}
\end{table}

Shared across every run: 4{,}000 / 8{,}000 / 16{,}000 total steps for 1 / 2 / 4 datasets, scaled so that the number of epochs per example is approximately constant across conditions; a linear warmup over the first 12.5\% of the budget followed by linear decay to zero; and seed 42. CRNN, Easter2.0, TrOCR-small standard and TrOCR-base standard are additionally retrained at seeds 43 and 44 on Karmapa8 and Drutsa, and Qwen3-VL at both seeds on Drutsa; their quantized variants inherit the seed of the checkpoint they quantize. Decoding is greedy, with TrOCR capped at 460 generated tokens and Qwen3-VL at 768.

CRNN, Easter2.0 and TrOCR share a character-level vocabulary built from the training split. Out-of-vocabulary test characters count as errors but no test split contains more than ten of them, at most 0.0013\% of test characters. Qwen3-VL keeps its pretrained vocabulary unchanged.

{\sloppy Where results are aggregated above individual configurations, the nine configurations are grouped into four families: \emph{CTC} (CRNN, Easter2.0), \emph{TrOCR-small} and \emph{TrOCR-base} (the standard, 8-bit and 4-bit variants of each), and \emph{Qwen3-VL}. The two TrOCR sizes are kept apart, differing by a factor of ten in parameters.\par}

All experiments ran on a single NVIDIA RTX 4050 Laptop GPU with 6~GB of VRAM, which restricts Qwen3-VL to QLoRA.

\subsection{Metrics}
\label{sec:evaluation}
\label{sec:eval-metrics}

All metrics are defined in \S\ref{sec:metrics} and computed directly on hypothesis and reference strings, following the \texttt{tibetan-wer} package\footnote{\url{https://pypi.org/project/tibetan-wer/}}. Rates are macro-averaged, each line weighted equally, except for the paired per-component comparisons of transfer, which are micro-averaged.

Leading and trailing whitespace is stripped from every reference and hypothesis before scoring. CER is computed over the codepoints of the stored strings, with no Unicode normalization or punctuation stripping. A syllable for SER is a maximal run of characters between delimiters: whitespace, the tsheg and tsheg bstar (U+0F0B--0F0C), the shad and the other head, clause and section marks (U+0F04--0F14 and U+0F34, U+0F3A--0F3D, U+0FD0--0FD4, U+0FD9--0FDA). The delimiters are not scored.

The three SWER variants use the segmenters of \S\ref{sec:metrics}. On the reference lines of the downstream subset, BoTok and BERT words average 1.3 syllables and Gemini words 2.9. Because of API cost, Gem-SWER is computed on a seeded random sample of 500 lines per condition, with the prompt of Appendix~\ref{app:prompt-seg}. On the mixed conditions the 500 lines are allocated across the components in proportion to their test-split sizes, and each component's lines are drawn from that component's single-domain sample. Before segmentation, strings whose \emph{character-type repetition ratio} --- length divided by number of distinct characters --- exceeds 10 are excluded, because API calls time out. A line is dropped if either its reference or hypothesis trips the filter.

\subsection{Error Concentration}
\label{sec:eval-operations}

\emph{Error concentration} is the share of error-free lines and the share of summed error contributed by the worst decile of lines, computed per cell over the per-line CER array and over the per-line SER array. It requires no segmenter.

\subsection{Statistical Analysis}
\label{sec:eval-stats}

Two systems compared on the same lines are tested with a paired bootstrap over test pages: pages are resampled 10{,}000 times, the difference is recomputed on the lines of the sampled pages, and it is significant when its 95\% interval excludes zero. No multiple-comparison correction is applied.

\label{sec:mixed-domain}
Transfer between the mixed and single-domain conditions is measured per component: the mixed-trained and single-domain-trained models are scored on the identical lines the split makes common to both (\S\ref{sec:split}) and the difference assessed by the paired bootstrap above.

A \emph{ranking reversal} is a pair of configurations that CER and every metric above the character level order in opposite directions. Gem-SWER is excluded from the search and from the paired transfer comparisons, being measured on a different sample (\S\ref{sec:eval-metrics}); where it differs from the other three, this is reported. The search is exhaustive over every pair of configurations in every condition, so a pair may differ in quantization, in size or in architecture. A reversal is \emph{significant} when the paired difference is significant on CER and, in the opposite direction, on at least two of SER, BoTok-SWER and BERT-SWER.

Correlations are quantified at two levels. Between per-line metric arrays we report Spearman $\rho$, computed within each model--condition cell separately. At the ranking level, models are ordered by each metric within each condition and the orderings compared with Kendall's $\tau$. The same procedure across the three SWER variants measures whether the segmenter changes the ordering.

\subsection{Downstream Task Evaluation}
\label{sec:downstream}

No gold English translation and no gold Tibetan entity annotation exist for these test sets, so absolute downstream quality cannot be measured. Following \citet{vanstrien2020impact}, a downstream system's output on clean text is taken as the ground truth for its output on OCR'd text. Downstream figures are therefore \emph{preservation} scores, measuring the effect of recognition error on the task, not the system's accuracy.

Evaluation uses the 500-line sample of each single-domain condition, the same lines as Gem-SWER, and all nine configurations (Table~\ref{tab:downstream}). Part-of-speech and entity preservation are F$_1$ over per-line items, repeated items counted separately, averaged over lines; translation preservation is corpus chrF \citep{popovic2015chrf} ($\beta = 2$, character 6-grams) and BLEU, both via sacreBLEU 2.6.0 \citep{post2018sacrebleu} with default settings, and sentence chrF and BLEURT \citep{sellam2020bleurt}, a learned metric rather than surface overlap, from the original BLEURT checkpoint, whose scores fall below zero for inadequate translations, each averaged over lines. Where the annotator returns no usable output for a hypothesis, the hypothesis is scored as empty output rather than dropped, so the failure counts against the system that produced the line. A line whose \emph{reference} the annotator could not process has no gold target and drops out. Downstream targets are set at round values of each measure (0.8 and 0.9 for part-of-speech F$_1$, 0.6 and 0.8 for entity F$_1$, 60 for chrF, 0.60 for BLEURT), chosen as reasonable levels of preservation for rough guidance; an error rate \emph{separates} a target when every cell meeting it has a lower rate than every cell missing it.

\begin{table}[h]
\caption{Downstream tasks. All are scored as preservation against reference-derived output. Entity labels are PER, LOC, ORG and WORK, the last for text titles.}
\label{tab:downstream}
\begin{tabular}{p{0.15\linewidth}p{0.29\linewidth}p{0.46\linewidth}}
\toprule
Task & Annotator & Measures reported \\
\midrule
Part-of-speech & \texttt{tibetan-bert-base-upos} & F$_1$ over (word, tag) pairs and over tags; tag-sequence error rate \\
Named entities & Gemini 2.5 Flash Lite & F$_1$ over (span, type) pairs and over types \\
Translation & Gemini 2.5 Flash Lite & corpus and sentence chrF, BLEU, BLEURT \\
\bottomrule
\end{tabular}
\Description{Table of the three downstream tasks, the annotator used for each, and the measures reported.}
\end{table}

\paragraph{Part-of-speech} Tagging uses \texttt{tibetan-bert-base-upos}, the model that also supplies the BERT-SWER segmentation, run locally: the word boundaries it induces are kept together with their UPOS labels, so each string becomes a sequence of (word, tag) pairs. The reference and every hypothesis are tagged this way, truncated to 800 characters and subject to the repetition filter of \S\ref{sec:eval-metrics}.

Entities and translation are annotated by Gemini 2.5 Flash Lite (\S\ref{sec:downstream-prior}). Requests are batched four lines at a time under the repetition filter of \S\ref{sec:eval-metrics}, each line truncated to 500 characters; prompts are in Appendix~\ref{app:prompt-mt-ner}.

\paragraph{Annotator coverage} For every string, whether usable output was returned is recorded, with reference lines as the control, separating strings excluded by the repetition filter from strings queried without parseable output. A failing request is retried, finally without the translation field (Appendix~\ref{app:prompt-ner-fallback}).

\section{Results}
\label{sec:results}

\subsection{Error Rates}
\label{sec:res-error-rates}

\begin{table}[t]
\caption{Macro-averaged error rates on the single-domain conditions, ordered by CER within each condition; lowest per condition in bold. The 8-bit variants, which differ from their standard counterparts by at most 0.014 in CER and 0.034 in SER, BoTok-SWER and BERT-SWER, are omitted.}
\label{tab:error-rates}
\small
\setlength{\tabcolsep}{4pt}
\begin{tabular}{lrrrrr}
\toprule
Configuration & CER & SER & BoTok & BERT & Gem \\
\midrule
\multicolumn{6}{l}{\emph{KhyentseWangpo}} \\
CRNN & \textbf{0.002} & \textbf{0.004} & \textbf{0.005} & \textbf{0.009} & \textbf{0.010} \\
Easter2.0 & 0.005 & 0.007 & 0.008 & 0.010 & 0.031 \\
TrOCR-base std & 0.007 & 0.017 & 0.023 & 0.054 & 0.109 \\
Qwen3-VL & 0.017 & 0.026 & 0.038 & 0.060 & 0.235 \\
TrOCR-base 4-bit & 0.028 & 0.078 & 0.119 & 0.194 & 0.507 \\
TrOCR-small std & 0.200 & 0.440 & 0.622 & 0.646 & 1.075 \\
TrOCR-small 4-bit & 0.653 & 0.989 & 1.168 & 1.110 & 1.365 \\
\addlinespace
\multicolumn{6}{l}{\emph{Karmapa8}} \\
CRNN & \textbf{0.013} & \textbf{0.031} & \textbf{0.043} & 0.105 & \textbf{0.290} \\
Easter2.0 & 0.014 & 0.032 & 0.045 & \textbf{0.092} & 0.358 \\
TrOCR-base std & 0.175 & 0.355 & 0.435 & 0.524 & 1.169 \\
TrOCR-base 4-bit & 0.353 & 0.633 & 0.733 & 0.756 & 1.494 \\
Qwen3-VL & 0.415 & 0.696 & 0.761 & 0.784 & 1.489 \\
TrOCR-small std & 0.551 & 0.910 & 0.958 & 0.975 & 1.614 \\
TrOCR-small 4-bit & 0.726 & 1.125 & 1.246 & 1.237 & 3.026 \\
\addlinespace
\multicolumn{6}{l}{\emph{Drutsa}} \\
CRNN & \textbf{0.047} & \textbf{0.073} & \textbf{0.104} & \textbf{0.175} & \textbf{0.622} \\
Easter2.0 & 0.078 & 0.133 & 0.187 & 0.234 & 0.849 \\
TrOCR-base std & 0.175 & 0.254 & 0.311 & 0.393 & 0.946 \\
Qwen3-VL & 0.251 & 0.434 & 0.517 & 0.580 & 1.243 \\
TrOCR-base 4-bit & 0.282 & 0.409 & 0.479 & 0.541 & 1.320 \\
TrOCR-small std & 0.645 & 0.944 & 0.980 & 0.958 & 1.491 \\
TrOCR-small 4-bit & 0.726 & 1.021 & 1.092 & 1.037 & 1.635 \\
\addlinespace
\multicolumn{6}{l}{\emph{Betsug}} \\
CRNN & \textbf{0.024} & \textbf{0.045} & \textbf{0.066} & 0.142 & 0.393 \\
Easter2.0 & 0.025 & 0.047 & 0.067 & \textbf{0.131} & \textbf{0.347} \\
TrOCR-base std & 0.109 & 0.171 & 0.221 & 0.321 & 0.660 \\
Qwen3-VL & 0.182 & 0.328 & 0.411 & 0.498 & 1.002 \\
TrOCR-base 4-bit & 0.212 & 0.344 & 0.425 & 0.512 & 0.843 \\
TrOCR-small std & 0.688 & 0.994 & 0.993 & 0.942 & 1.320 \\
TrOCR-small 4-bit & 0.691 & 1.151 & 1.114 & 1.061 & 1.420 \\
\bottomrule
\end{tabular}
\Description{Table of macro-averaged CER, SER, BoTok-SWER, BERT-SWER and Gem-SWER for each configuration, grouped by condition.}
\end{table}

CRNN records the lowest CER in all four single-domain conditions (Table~\ref{tab:error-rates}) and in two of the three mixed ones, Easter2.0 in the third. On Karmapa8 and Betsug CRNN leads Easter2.0 by 0.001 CER, and on Karmapa8 the two exchange places at one training seed (\S\ref{sec:res-ranking}).

\subsection{Agreement Between Metric Levels}
\label{sec:res-agreement}

\subsubsection{Model rankings}
\label{sec:res-ranking}

\begin{table}[t]
\caption{Kendall $\tau$ between model rankings of the configurations: under CER against under each syllable- and word-level metric, and under each pair of segmenters.}
\label{tab:ranking-tau}
\footnotesize
\setlength{\tabcolsep}{4pt}
\begin{tabular}{lrrrrrrrr}
\toprule
& & \multicolumn{4}{c}{CER against} & \multicolumn{3}{c}{Segmenter pairs} \\
\cmidrule(lr){3-6}\cmidrule(lr){7-9}
Condition & $n$ & SER & BoTok & BERT & Gem & BoTok$\sim$BERT & BoTok$\sim$Gem & BERT$\sim$Gem \\
\midrule
KhyentseWangpo & 9 & 1.000 & 1.000 & 1.000 & 1.000 & 1.000 & 1.000 & 1.000 \\
Karmapa8 & 9 & 0.944 & 1.000 & 0.944 & 0.889 & 0.944 & 0.889 & 0.833 \\
Drutsa & 9 & 0.944 & 0.889 & 0.944 & 0.944 & 0.944 & 0.833 & 0.889 \\
Betsug & 9 & 0.944 & 0.944 & 0.889 & 0.833 & 0.944 & 0.889 & 0.944 \\
Drutsa+Betsug & 8 & 1.000 & 1.000 & 1.000 & 1.000 & 1.000 & 1.000 & 1.000 \\
KhyentseWangpo+Karmapa8 & 8 & 0.929 & 0.929 & 0.929 & 0.929 & 1.000 & 1.000 & 1.000 \\
All four & 8 & 0.929 & 0.929 & 0.929 & 0.929 & 1.000 & 1.000 & 1.000 \\
\bottomrule
\end{tabular}
\Description{Table of Kendall tau between model rankings under CER and under each syllable- and word-level metric, and between the rankings produced by each pair of segmenters, per condition.}
\end{table}

Kendall $\tau$ between the CER ranking and each syllable- or word-level ranking ranges from 0.833 to 1.000 across the 28 condition $\times$ metric cells of Table~\ref{tab:ranking-tau}. The three segmenters produce rankings that agree at a minimum $\tau$ of 0.833, a median of 1.000 and a maximum of 1.000. The minimum is shared by the Karmapa8 pairing of BERT-SWER and Gem-SWER and the Drutsa pairing of BoTok-SWER and Gem-SWER; BoTok and BERT agree at 0.944--1.000 in every condition, and SER ranks systems as BoTok-SWER and BERT-SWER do at 0.889--1.000.

At two further training seeds for CRNN, Easter2.0, TrOCR-small standard and TrOCR-base standard, $\tau$ between seeds is 1.000 on Drutsa under CER, SER and BoTok-SWER. On Karmapa8 it is 0.667 for two of the three seed pairs under all three metrics: CRNN and Easter2.0, whose CERs differ by 0.001, exchange places at one seed. Per-configuration CER varies across seeds by 0.000--0.009 for the CTC models, 0.021--0.072 for the TrOCR configurations and 0.007 for Qwen3-VL.

\subsubsection{Per-line agreement and the error rate}
\label{sec:res-perline}

\begin{figure}[t]
\centering
\includegraphics[width=0.82\linewidth]{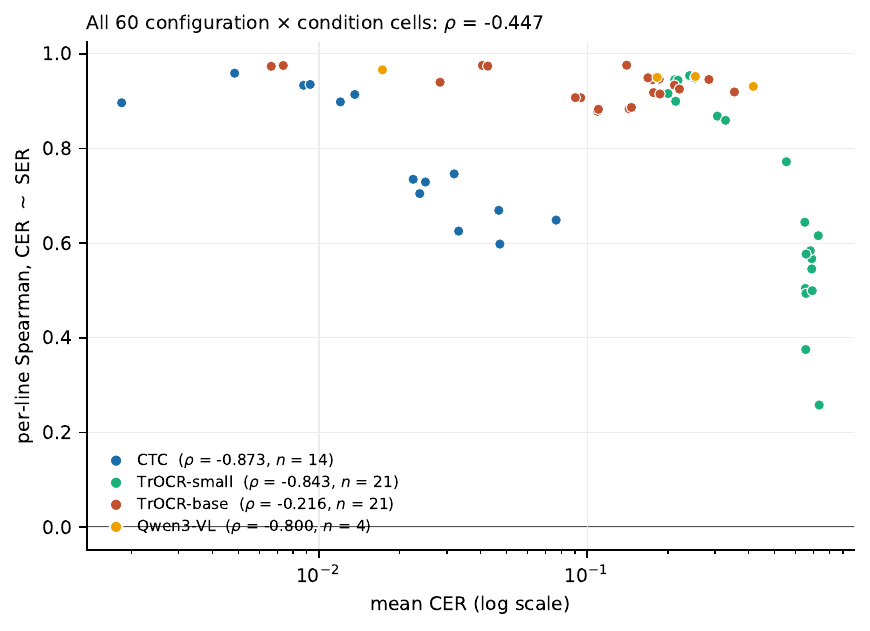}
\caption{Per-line Spearman correlation between CER and SER against the mean CER
of the same cell, for all 60 configuration $\times$ condition cells. Legend entries give the correlation
within each family. Agreement falls with mean CER and, at comparable mean CER, is lower for the
CTC cells on the conditions containing handwriting, much of whose character error
falls on whitespace.}
\label{fig:agreement-vs-cer}
\Description{Scatter plot of per-line CER-SER correlation against mean CER on a
logarithmic axis, coloured by architecture family. Agreement falls with mean CER,
to between 0.26 and 0.77 for the TrOCR-small cells above mean CER 0.5. The CTC cells
on the handwriting conditions lie at 0.60--0.75 at mean CER 0.02--0.08, below the
TrOCR-base cells at comparable error rates.}
\end{figure}

Across the 60 configuration $\times$ condition cells, a cell's mean CER and its per-line CER$\sim$SER agreement correlate at $\rho = -0.447$, and within each family separately at CTC $-0.873$ (14 cells), TrOCR-small $-0.843$ (21), TrOCR-base $-0.216$ (21) and Qwen3-VL $-0.800$ (4). At comparable error rates, families differ (Figure~\ref{fig:agreement-vs-cer}): the CTC cells on the conditions containing handwriting record agreement of 0.598--0.746 at mean CER 0.022--0.077, and the three TrOCR-base cells in that range 0.940--0.975. A median of 30.9\% of a CTC cell's character edits fall on whitespace, which SER does not score (\S\ref{sec:eval-metrics}), against 11.3\% for TrOCR-base, 4.7\% for TrOCR-small and 1.9\% for Qwen3-VL. The correlation between mean CER and per-line agreement is $-0.588$ for CER$\sim$BoTok-SWER, $-0.500$ for CER$\sim$BERT-SWER, $-0.693$ for SER$\sim$BoTok-SWER, $-0.346$ for SER$\sim$BERT-SWER and $-0.318$ for BoTok-SWER$\sim$BERT-SWER. Agreement among SER, BoTok-SWER and BERT-SWER is positive in all 60 cells, with a minimum of 0.256.

\subsubsection{Reversals and sign disagreements}
\label{sec:res-disagreements}

One configuration pair is a significant reversal (\S\ref{sec:eval-stats}). On Drutsa, CER ranks Qwen3-VL ahead of TrOCR-base 4-bit by 0.031 $[0.007, 0.058]$. BoTok-SWER and BERT-SWER rank TrOCR-base 4-bit ahead by 0.037 and 0.039, and SER by 0.024, whose interval includes zero. Across the nine pairings of the two configurations at seeds 42--44, CER and the three higher-level rates order the pair oppositely in all nine. On the downstream sample, sentence chrF favours the higher-level ordering by 5.37 $[2.53, 8.33]$, and part-of-speech and entity F$_1$ do not differ.

Joint training was compared with single-domain training per component on the 64 configuration $\times$ component cells, each a component's full test split of 1{,}355--3{,}244 lines (\S\ref{sec:eval-stats}). CER records a significant reduction in 48 cells and a significant increase in 11; SER 46 and 13, BoTok-SWER 46 and 10, BERT-SWER 43 and 15. In ten cells CER and a syllable- or word-level metric record significant changes of opposite sign. Eight are TrOCR-small 4-bit cells, in each of which CER records the reduction. The other two are TrOCR-base standard and 8-bit on the Betsug component of Drutsa+Betsug, in which CER records the increase. No CTC cell disagrees on sign. By pair, CER disagrees in sign with SER in seven cells, five of them TrOCR-small 4-bit, with BoTok-SWER in five and with BERT-SWER in eight; SER, BoTok-SWER and BERT-SWER disagree with one another in two or three cells per pair.

\subsection{Error Concentration}
\label{sec:res-structure}
\label{sec:res-concentration}

\begin{table}[t]
\caption{Concentration of per-line error, median over each family's cells. \emph{Clean} is the share of lines with no error and \emph{worst decile} the share of all error falling in the worst tenth of lines.}
\label{tab:concentration}
\small
\setlength{\tabcolsep}{4pt}
\begin{tabular}{lrrrrr}
\toprule
& & \multicolumn{2}{c}{Per-line CER} & \multicolumn{2}{c}{Per-line SER} \\
\cmidrule(lr){3-4}\cmidrule(lr){5-6}
Family & $n$ & Clean \% & Worst decile \% & Clean \% & Worst decile \% \\
\midrule
CTC & 8 & 28.2 & 49.8 & 38.2 & 46.0 \\
TrOCR-small & 12 & 0.1 & 14.9 & 0.1 & 13.1 \\
TrOCR-base & 12 & 7.9 & 42.9 & 12.5 & 30.1 \\
Qwen3-VL & 4 & 1.9 & 26.8 & 2.8 & 22.7 \\
\bottomrule
\end{tabular}
\Description{Table of error-free line share and worst-decile error share per architecture family, for per-line CER and per-line SER.}
\end{table}

\begin{figure}[t]
\centering
\includegraphics[width=0.7\linewidth]{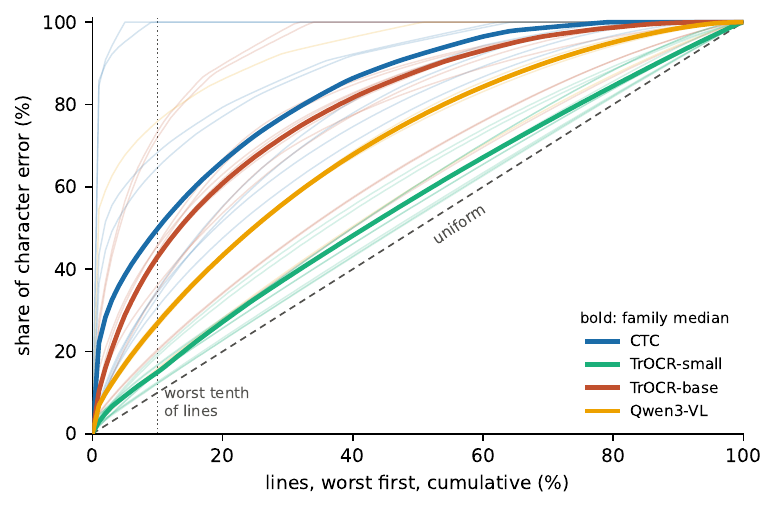}
\caption{Concentration of character error across lines, for the 36 single-domain
cells. Each curve is the cumulative share of summed per-line CER
against the cumulative share of lines, worst first; bold curves are family
medians and the dashed diagonal is a uniform distribution. A curve's value at the
dotted line is its worst-decile share (Table~\ref{tab:concentration}). CTC and
TrOCR-base concentrate error in few lines; TrOCR-small spreads it almost uniformly.}
\label{fig:concentration}
\Description{Lorenz curves for 36 cells, faint, with one bold median curve per
architecture family. The CTC and TrOCR-base medians rise steeply, reaching about 50
and 43 percent of error at 10 percent of lines; the TrOCR-small median lies close to the
uniform diagonal, with Qwen3-VL between them.}
\end{figure}

Concentration differs by family (Table~\ref{tab:concentration}, Figure~\ref{fig:concentration}). CTC places 49.8\% of its character error in the worst tenth of lines and leaves 28.2\% of lines clean, TrOCR-base 42.9\% and 7.9\%, and Qwen3-VL 26.8\% and 1.9\%. TrOCR-small spreads error almost uniformly, 14.9\% in the worst decile, and leaves 0.1\% clean. The ordering is the same under SER, with every worst-decile share lower; the CTC clean share rises to 38.2\%, since lines whose only errors fall on whitespace or punctuation are clean under SER. By domain, CTC leaves 90.9--95.0\% of lines unchanged on KhyentseWangpo and 35.7--37.6\% on Karmapa8; on Drutsa and Betsug every configuration has at least 79.3\% of lines differing from the reference.

\subsection{Downstream Consequences}
\label{sec:res-downstream}

\subsubsection{Downstream preservation}
\label{sec:res-downstream-scores}

\begin{table}[t]
\caption{Downstream preservation, median over each family's configurations, on the 500-line seeded random subset of each single-domain condition. POS F$_1$ is over (word, tag) pairs, NER F$_1$ over (surface form, type) pairs; chrF and BLEURT score the translation against the one produced from the reference. CER and SER are family medians of the full test-set rates. Bold: the downstream target of \S\ref{sec:res-downstream-scores} is met (POS and NER F$_1$ 0.8, chrF 60).}
\label{tab:downstream-scores}
\small
\setlength{\tabcolsep}{4pt}
\begin{tabular}{lrrrrrrr}
\toprule
Family & $n$ & CER & SER & POS F$_1$ & NER F$_1$ & chrF & BLEURT \\
\midrule
\multicolumn{8}{l}{\emph{KhyentseWangpo}} \\
CTC & 2 & 0.003 & 0.005 & \textbf{0.995} & \textbf{0.946} & \textbf{96.4} & +0.835 \\
TrOCR-small & 3 & 0.214 & 0.465 & 0.451 & 0.241 & 28.8 & -0.738 \\
TrOCR-base & 3 & 0.007 & 0.019 & \textbf{0.952} & 0.794 & \textbf{83.1} & +0.517 \\
Qwen3-VL & 1 & 0.017 & 0.026 & \textbf{0.955} & 0.688 & \textbf{75.3} & +0.339 \\
\addlinespace
\multicolumn{8}{l}{\emph{Karmapa8}} \\
CTC & 2 & 0.014 & 0.032 & \textbf{0.919} & 0.704 & \textbf{70.8} & +0.250 \\
TrOCR-small & 3 & 0.552 & 0.910 & 0.271 & 0.322 & 19.8 & -1.047 \\
TrOCR-base & 3 & 0.185 & 0.372 & 0.555 & 0.429 & 37.1 & -0.637 \\
Qwen3-VL & 1 & 0.415 & 0.696 & 0.371 & 0.315 & 26.6 & -0.932 \\
\addlinespace
\multicolumn{8}{l}{\emph{Drutsa}} \\
CTC & 2 & 0.063 & 0.103 & \textbf{0.818} & 0.619 & 52.6 & -0.153 \\
TrOCR-small & 3 & 0.648 & 0.944 & 0.177 & 0.438 & 15.2 & -1.181 \\
TrOCR-base & 3 & 0.185 & 0.256 & 0.662 & 0.568 & 41.9 & -0.332 \\
Qwen3-VL & 1 & 0.251 & 0.434 & 0.518 & 0.501 & 35.4 & -0.713 \\
\addlinespace
\multicolumn{8}{l}{\emph{Betsug}} \\
CTC & 2 & 0.024 & 0.046 & \textbf{0.888} & 0.594 & \textbf{62.7} & +0.036 \\
TrOCR-small & 3 & 0.688 & 0.998 & 0.194 & 0.180 & 16.8 & -1.211 \\
TrOCR-base & 3 & 0.110 & 0.176 & 0.722 & 0.488 & 49.4 & -0.324 \\
Qwen3-VL & 1 & 0.182 & 0.328 & 0.598 & 0.397 & 39.4 & -0.602 \\
\bottomrule
\end{tabular}
\Description{Table of part-of-speech, named-entity and translation preservation by architecture family and condition, beside the corresponding character and syllable error rates.}
\end{table}

\begin{figure}[t]
\centering
\includegraphics[width=\linewidth]{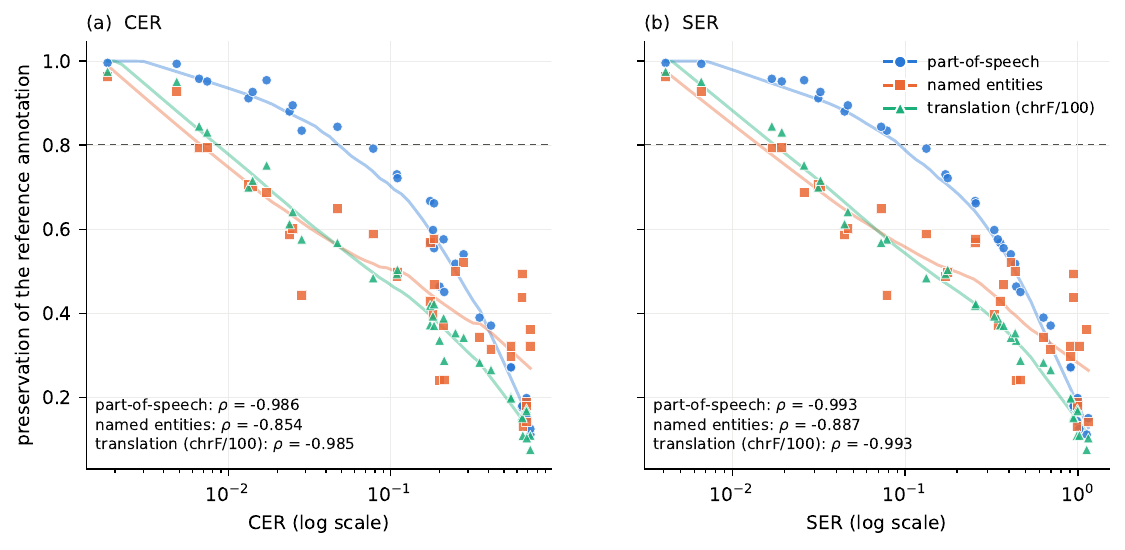}
\caption{Preservation of the reference annotation on three downstream tasks,
against (a) CER and (b) SER on logarithmic axes, over the 36 configuration
$\times$ condition cells for which downstream measures were computed. Curves are
LOWESS fits on the logarithm of the rate, one per task; dashed lines mark the
targets of \S\ref{sec:res-downstream-scores} (0.8 for part-of-speech and entity
preservation, 60 for chrF). Translation chrF is divided by 100 to share the axis.
SER's association
with each measure is within 0.005 of the strongest of the five rates
(Table~\ref{tab:downstream-prediction}).}
\label{fig:downstream-scatter}
\Description{Two scatter panels of downstream preservation against error rate on
logarithmic axes, with a fitted curve per task and dashed lines at 0.8 and 0.6.
Named-entity preservation lies below part-of-speech preservation in 30 of the 36 cells.}
\end{figure}

Preservation declines as the error rate rises on all three tasks --- part-of-speech tagging, named-entity recognition and translation --- and declines furthest on named-entity recognition (Table~\ref{tab:downstream-scores}, Figure~\ref{fig:downstream-scatter}). The CTC models on typeset text, at a median CER of 0.003, score 0.995 on part-of-speech preservation, 0.946 on entity preservation and 96.4 chrF. The same models on Betsug, at CER 0.024--0.025, score 0.880--0.895 on part-of-speech preservation while entity preservation falls to 0.587--0.602. At the other end, TrOCR-small on the two handwriting conditions reaches CER 0.65--0.73, with entity preservation of 0.142--0.494 and BLEURT of $-1.39$ to $-1.18$. Entity preservation is below part-of-speech preservation in 30 of the 36 measurements, by a median of 0.09.

Downstream targets are set at round values (\S\ref{sec:downstream}). Every target except entity preservation of 0.6 is separated by CER, SER, BoTok-SWER and BERT-SWER. The seven cells reaching part-of-speech preservation of 0.9 are set apart from the remaining 29 at CER 0.017--0.024 and at SER 0.032--0.045. Entity preservation of 0.8 is reached only at SER 0.007 or below (CER 0.005--0.007), chrF 60 at SER 0.047 or below (CER 0.025--0.028), and part-of-speech preservation of 0.8 at SER 0.078 or below (CER 0.047--0.078). BLEURT 0.60 is reached only at SER 0.007 or below (CER 0.005--0.007). No rate separates the nine cells reaching entity preservation of 0.6 from those missing it: Drutsa CRNN reaches it at SER 0.073 (CER 0.047), and Betsug CRNN misses it at SER 0.045 (CER 0.024).

\subsubsection{Predictors of downstream degradation}
\label{sec:res-downstream-pred}

\begin{table}[t]
\caption{Spearman correlation of each error rate with each downstream preservation score over the 36 cells, strongest per row in bold. SER is within 0.005 of the strongest rate on every measure.}
\label{tab:downstream-prediction}
\footnotesize
\setlength{\tabcolsep}{4pt}
\begin{tabular}{lrrrrr}
\toprule
Downstream measure & CER & SER & BoTok & BERT & Gem \\
\midrule
POS (word, tag) F$_1$ & -0.986 & -0.993 & \textbf{-0.996} & -0.992 & -0.956 \\
POS tag-only F$_1$ & \textbf{-0.992} & -0.991 & -0.988 & -0.982 & -0.960 \\
POS tag-sequence ER & 0.982 & 0.991 & 0.993 & \textbf{0.996} & 0.963 \\
NER (form, type) F$_1$ & -0.854 & -0.887 & \textbf{-0.889} & -0.877 & -0.790 \\
NER type-only F$_1$ & -0.932 & -0.955 & \textbf{-0.956} & -0.946 & -0.886 \\
MT chrF & -0.985 & -0.993 & \textbf{-0.995} & -0.992 & -0.956 \\
MT sentence chrF & -0.979 & -0.991 & \textbf{-0.993} & -0.990 & -0.950 \\
MT BLEU & -0.985 & -0.994 & \textbf{-0.997} & -0.990 & -0.947 \\
MT BLEURT & -0.981 & -0.994 & \textbf{-0.995} & -0.988 & -0.944 \\
\bottomrule
\end{tabular}
\Description{Table of Spearman correlations between each error rate and each downstream preservation measure.}
\end{table}

The three part-of-speech measures are tagged locally and do not depend on an LLM annotator. On every measure CER, SER, BoTok-SWER and BERT-SWER correlate with preservation within 0.035 of one another (Table~\ref{tab:downstream-prediction}; Figure~\ref{fig:downstream-scatter} plots CER and SER against three of the measures). Gem-SWER, computed on the same 500 lines but with lines dropped by its repetition filter, is the weakest rate on all nine measures.

\subsubsection{Annotator coverage}
\label{sec:res-annotator}

The entity and translation scores reported above come from Gemini 2.5 Flash Lite; part-of-speech preservation is tagged locally and does not depend on it. A line the annotator cannot process is scored as empty output (\S\ref{sec:downstream}), so every measure covers at least 493 of the 500 lines per cell; the lines that drop out are those whose \emph{reference} it could not process. The annotator returns no usable output, an empty or unparseable response, for up to 1.4\% of a cell's lines, at a rate that correlates with CER at $\rho = +0.622$.

\section{Discussion}
\label{sec:discussion}

\subsection{Ranking Agreement}
\label{sec:disc-rankings}

In general, rankings of systems agree across the metric levels. Kendall $\tau$ between the CER ranking and each syllable- or word-level ranking never falls below 0.833 (\S\ref{sec:res-ranking}). This is consistent with \citet{vesalainen2026errorpatterns}, but contrasts with the ranking reversals seen elsewhere \citep{agbetimessan2026ssmocr,le2025vietnamesedar,moore2026tibetanasr}. We find only one reversal, which holds across training seeds. On it, downstream translation favours the ordering given by the segmented word error rates (\S\ref{sec:res-disagreements}).

Rankings computed over the three segmenters agree at a median pairwise $\tau$ of 1.000 (\S\ref{sec:res-ranking}). Which segmenter is used does not change which system a word-level rate judges better, as \citet{moore2026tibetanasr} found for automatic speech recognition.

Line by line, CER and SER diverge for two reasons. Across cells, per-line CER$\sim$SER agreement falls with mean CER at $\rho = -0.447$ (\S\ref{sec:res-perline}). At a given error rate, it also depends on how much of a system's character error falls on whitespace, which CER counts and SER does not: the CTC models, with a median of 30.9\% of character edits on whitespace, agree at 0.598--0.746 on the conditions containing handwriting, where TrOCR-base cells at comparable CER agree at 0.940--0.975. A CER-only evaluation of such a system therefore ranks its lines partly by whitespace errors, which SER does not score. Agreement among SER, BoTok-SWER and BERT-SWER is positive in every cell. \citet{moore2026tibetanasr} report the CER--SER correlation falling under parameter-efficient fine-tuning; here it falls with the error rate within the CTC and TrOCR-small families.

\subsection{Disagreement Between the Rates}
\label{sec:disc-interventions}

On a decision about a single system, CER can give the opposite answer from the rates above it. Comparing joint with single-domain training on identical lines, CER and a syllable- or word-level rate record significant changes of opposite sign in ten of the 64 cells, eight of them TrOCR-small 4-bit cells, where CER records an improvement (\S\ref{sec:res-disagreements}).

Conveniently, SER requires no segmenter, but ranks systems as BoTok-SWER and BERT-SWER do at Kendall $\tau \geq 0.889$ (\S\ref{sec:res-ranking}, \S\ref{sec:eval-metrics}); the words of those two segmenters average 1.3 syllables. The same reasoning led \citet{naing2015myanmarlvcsr} to a syllable error rate for Myanmar speech recognition, where the definition of a word is ambiguous.

\subsection{Error Rate and Downstream Use}
\label{sec:disc-downstream}

As rough guidance, with arbitrary but reasonable targets, part-of-speech preservation of 0.8 is reached only at SER 0.078 or below and translation chrF of 60 at 0.047 or below; the corresponding CER bands are 0.047--0.078 and 0.025--0.028 (\S\ref{sec:res-downstream}). No rate separates the systems that reach entity preservation of 0.6 from those that miss it. The 5\% CER that \citet{griffiths2024htr} sets for text mining falls within the tagging band; the 10\% described there as efficient for automatic transcription lies above it.

Across the three tasks, the SER needed ranges from 0.007 for entity preservation of 0.8 to 0.078 for part-of-speech preservation of 0.8, so no single acceptable rate serves every downstream use of Tibetan OCR at either the character or the syllable level. Entity preservation degrades fastest of the three tasks and is the hardest to predict, at $\rho = -0.854$ for CER and $-0.887$ for SER (Table~\ref{tab:downstream-prediction}). \citet{hamdi2019nerocr} also report entity extraction degrading sharply under OCR error, and \citet{jaud2025beyondcer}, using simulated OCR noise, find the loss concentrated in errors that fall on entity spans.

Entity extraction and translation are performed by an LLM, Gemini 2.5 Flash Lite, which fails to return usable output more often as its input degrades (\S\ref{sec:res-annotator}). Unreliable behaviour of language and translation models on noisy OCR text, on perturbed source input and in low-resource translation has been reported before \citep{piryani2025multiocrqa,raunak2021hallucinations,guerreiro2023hallucinations}. Failures are scored as empty output, so they count against the score. An evaluation that places an LLM annotator downstream of OCR ought thus to report the annotator's coverage and how its failures are scored.

\section{Limitations and Future Work}
\label{sec:future}

\subsection{Limitations}
\label{sec:limitations}

Splits are grouped by page, not by text or scribe. Other pages of a test page's text, and other lines in its scribe's hand, can occur in training, so the error rates reported may be lower than those on unseen texts or hands, particularly on handwriting.

Training settings differ by architecture: batch size, learning rate and the number of passes over the training data were each set by hardware constraints and training stability (Table~\ref{tab:arch-hparams}). Differences between architecture families therefore include differences in optimization.

Part-of-speech tagging and BERT-SWER use the same model, so the two are not independent. Gem-SWER, entity extraction and translation use a hosted model that cannot be pinned to a version.

The downstream figures are preservation against a fixed annotator's output on the reference rather than accuracy, as no gold Tibetan entity annotation or English translation exists for these test sets (\S\ref{sec:downstream}). They cover the nine configurations on a 500-line random sample of each single-domain test split (\S\ref{sec:eval-metrics}). Gem-SWER is computed on the same lines. The reliability of LLMs as evaluators varies across models and tasks \citep{bavaresco2025llmjudges}, particularly in low-resource and multilingual settings \citep{fu2025multilingualjudge,dogruoz2026judgelowresource}, and the same may hold for LLM annotators. A single annotator is used, so its errors are not separated from those caused by recognition. About 4\% of entity spans, on references and hypotheses alike, are returned as English renderings that cannot match a Tibetan span, so the entity bands in particular are relative quantities. The downstream targets are round values chosen for rough guidance.

\subsection{Future Work}
\label{sec:futurework}

Future work should extend this analysis to other scripts that leave word boundaries unmarked, such as other Tibetic languages (e.g. Dzongkha) or the Brahmic abugidas of mainland Southeast Asia (Thai, Lao, Khmer and Burmese).

Evaluation should also be extended to the full pipeline, with layout detection and line segmentation before recognition, where the page-level measures of \citet{bourne2026cev} apply; this study assumes correct line segmentation (\S\ref{sec:metrics}). Scoring one recognizer on gold and on predicted line crops would separate segmentation error from recognition error, particularly on handwriting.

Post-editing effort should be measured directly and related to per-line error concentration. Correction time has been measured for OCR \citep{rijhwani2023kwakwala} and transcription time with ASR output visible for speech \citep{sperber2017transcribing}, and for machine translation edit distance to the post-edited output has been compared with perceived effort \citep{koponen2012postediting}.
\section{Conclusion}
\label{sec:conclusion}

This study reports syllable- and word-level error rates for Tibetan OCR, to our knowledge for the first time. Nine configurations from three architecture classes, CTC recognizers, TrOCR encoder--decoders with and without post-training quantization and a vision-language model adapted with QLoRA, were evaluated on seven conditions drawn from typeset, woodblock and two handwritten script styles. Each was scored under CER, SER and three segmented word error rates on page-grouped held-out splits with bootstrap intervals. The rates were compared to one another, to the distribution of error over lines, and to the preservation of part-of-speech tags, named entities and translations computed from the output.

Tibetan OCR evaluations should report SER alongside CER. SER requires no segmenter, lexicon or model. CER broadly agrees with the higher-level rates on rankings, with one reversal, on which translation quality favours the ordering given by the segmented rates (\S\ref{sec:disc-rankings}). On decisions about a single system, such as whether joint training helped, CER and SER can disagree in sign (\S\ref{sec:disc-interventions}).

Where an LLM is used to annotate OCR output for downstream evaluation, we recommend that its coverage be reported and its failures scored rather than dropped. Gemini 2.5 Flash Lite returned no usable output for up to 1.4\% of lines, more often as its input degraded (\S\ref{sec:disc-downstream}).

As rough guidance for downstream use, the SER needed ranges from 0.078 or below for part-of-speech preservation of 0.8 to 0.007 or below for entity preservation of 0.8, so no single acceptable rate serves every downstream use. No rate separates the systems that reach entity preservation of 0.6 from those that do not (\S\ref{sec:disc-downstream}).

\begin{acks}
{\sloppy We thank the Buddhist Digital Resource Center (BDRC) and WeBuddhist (formerly OpenPecha) for making the KhyentseWangpo, Karmapa8, Drutsa and Betsug datasets publicly available.\par}

Claude (Anthropic), used through Claude Code, assisted with writing code for training and analysis, generating figures and tables, and drafting and revising the text.
\end{acks}

\section*{Data Availability}
\label{sec:availability}

{\sloppy The four datasets are public Hugging Face repositories (Table~\ref{tab:dataset-stats}). TrOCR is fine-tuned from the public checkpoints \texttt{microsoft/\allowbreak trocr-small-stage1} and \texttt{microsoft/\allowbreak trocr-base-stage1}, and Qwen3-VL from \texttt{Qwen/\allowbreak \mbox{Qwen3-VL-2B-Instruct}}; CRNN and Easter2.0 are trained from random initialization. BERT-SWER segmentation and part-of-speech tagging use the public \texttt{KoichiYasuoka/\allowbreak tibetan-bert-base-upos}. CER, SER and the three SWER variants are computed with the public \texttt{tibetan-wer} package (\url{https://pypi.org/project/tibetan-wer/}). All prompts are given in Appendix~\ref{app:prompts}. Training and evaluation code, trained checkpoints and recognizer outputs are available from the authors on reasonable request.\par}

\bibliographystyle{ACM-Reference-Format}
\bibliography{references}

\appendix
\section{Prompts}
\label{app:prompts}

All prompts are reproduced verbatim. Placeholders in braces are filled at call
time. The Gemini prompts are issued at temperature 0.0 under a constrained JSON
response schema.

\subsection{Line transcription (Qwen3-VL)}
\label{app:prompt-ocr}

Issued in the model's native chat format alongside the line image
(\S\ref{sec:qwen}):

\begin{quote}\ttfamily\small\raggedright
Transcribe the Tibetan text in this line image. Output only the text.
\end{quote}

\subsection{Word segmentation for Gem-SWER}
\label{app:prompt-seg}

\begin{quote}\ttfamily\small\raggedright
Segment the following Tibetan text into words. Output ONLY the segmented text
with words separated by a pipe character (|). Keep the original Tibetan
characters and tsek marks (\tib{་}) intact within each word. Do not add any
explanation or punctuation beyond the pipe separators. Text: \{text\}
\end{quote}

\subsection{Translation and entity extraction}
\label{app:prompt-mt-ner}

One call returns both fields for a batch of four numbered lines
(\S\ref{sec:downstream}):

\begin{quote}\ttfamily\small\raggedright
You are processing lines of Tibetan text, some of which are noisy OCR output
and may be garbled or nonsensical.

For EACH numbered line below, return:

\ \ "translation": a literal English translation, AT MOST 40 WORDS. Never
repeat a word, phrase, whitespace or punctuation mark to pad the output. If the
line is too garbled to translate, give the best partial translation you can in a
few words; return "" only if nothing at all is recoverable.

\ \ "entities": named entities appearing in that line, each as \{"text":
<the Tibetan substring>, "type": one of "PER" (person, including teachers,
lineage holders and deities), "LOC" (place), "ORG" (monastery, school,
institution), "WORK" (title of a text or scripture)\}. Return an empty list if
there are none.

Do not merge or reorder lines. Return exactly one result object per input line,
in order.

Lines:

\{lines\}
\end{quote}

\subsection{Entity-only fallback}
\label{app:prompt-ner-fallback}

Issued to a single line when the combined request above returns no parseable
output. It is the prompt of Appendix~\ref{app:prompt-mt-ner} with the
\texttt{translation} field and its instruction removed, so only the entity list
is requested; this recovers entity coverage on degraded lines whose translation
runs to the output-token cap.

\end{document}